\documentclass[letterpaper, 10 pt, conference]{ieeeconf}  

\IEEEoverridecommandlockouts                              

\usepackage{epsfig} 
\usepackage{mathptmx} 
\usepackage{times} 
\usepackage{amsmath} 
\usepackage{amssymb}  

\usepackage{booktabs}
\usepackage{cite}
\usepackage{comment}

\let\labelindent\relax
\usepackage{enumitem}
\usepackage{float}
\usepackage{graphicx}
\usepackage{hyperref}
\usepackage[utf8]{inputenc}
\usepackage{multirow}
\usepackage[caption=false]{subfig}

\title{\LARGE \bf
Unpacking Human Teachers' Intentions for\\ Natural Interactive Task Learning}

\author{Preeti Ramaraj$^{1}$, Charles L. Ortiz, Jr.$^{2}$ and Shiwali Mohan$^{2}$
\thanks{$^{1}$Computer Science and Engineering, University of Michigan, Ann Arbor, Michigan 48109, USA {\tt\small preetir@umich.edu}}%
\thanks{$^{2}$Palo Alto Research Center, Palo Alto, California 94304, USA {\tt\small cortiz@parc.com, smohan@parc.com}}%
}

\begin{document}


\maketitle
\thispagestyle{empty}
\pagestyle{empty}

\begin{abstract}
Interactive Task Learning (ITL) is an emerging research agenda that studies the design of complex intelligent robots that can acquire new knowledge through natural human teacher-robot learner interactions. ITL methods are particularly useful for designing intelligent robots whose behavior can be adapted by humans collaborating with them. Various research communities are contributing methods for ITL and a large subset of this research is \emph{robot-centered} with a focus on developing algorithms that can learn online, quickly. This paper studies the ITL problem from a \emph{human-centered} perspective to provide guidance for robot design so that human teachers can naturally teach ITL robots. In this paper, we present 1) a qualitative bidirectional analysis of an interactive teaching study (N=10) through which we characterize various aspects of actions intended and executed by human teachers when teaching a robot; 2) an in-depth discussion of the teaching approach employed by two participants to understand the need for personal adaptation to individual teaching styles; and 3) requirements for ITL robot design based on our analyses and informed by a computational theory of collaborative interactions, SharedPlans.
\end{abstract}

\section{Introduction}
\vspace{-0.2cm}
We envision a future where robots can help people with a myriad set of tasks in dynamic environments such as homes, offices, shopping centers, and warehouses. However, the eventual set of tasks that a robot will be requested to perform when deployed will often be more diverse than can be planned for at design time. An attractive solution is to enable people to teach robots new tasks and relevant information about their environments on the fly, which is the focus of the approach called Interactive Task Learning (ITL) \cite{laird2017interactive}. 
ITL relies on the fact that people naturally engage in interactive teaching and learning, and therefore can apply those skills to teach a robot as well. A natural starting point, then, is to better understand how humans teach. 

Human teachers engage in a variety of interactive behaviors aimed at structuring the learner's learning experience. An important component  is the teacher's mental model \cite{norman1983some} of the learner: that is, understanding what the learner knows and does not know and how a learner can apply its knowledge to perform various tasks in the environment. A teacher may apply various interactive strategies to develop this understanding about the learner. For example, a human parent (teacher) may ask a child (learner) to demonstrate a skill (\emph{fold the towel}), to identify concepts (\emph{where is your head?}), to instantiate a concept (\emph{can you show me what angry looks like?}), or compare objects (\emph{is an orca bigger than a beluga?}). Responses to such requests aid the teacher in adapting instruction that is more suitable to the learner's needs. Without reasonable estimates of a learner’s capabilities, it is challenging for a human teacher to teach effectively. 
A learner must also be able to understand what the teacher intends with each instruction and respond appropriately while absorbing new information as it is presented.


Our research goal, therefore, is to build ITL robots that can engage in teaching interactions that are natural for humans. For this, we look to plan-based theories of dialogue \cite{grosz-kraus,ortiz-dial} that posit that exchanges between participants are intentional and reflect the evolution of their mental states during collaboration. 
Previous work \cite{breazeal2004teaching,Mohan2012,mohseni2015} has leveraged these theories to develop computational models that can manage ITL robot interactions. While those approaches enable flexible and mixed-initiative interaction, they are \emph{robot-centered} and are largely driven by the robot’s learning needs. 

In this paper, we take a \emph{human-centered} view, and focus on uncovering the structure in human teaching and the diversity of information it encompasses to support effective interaction management in ITL robots. 
We describe a human participant study (N=10) in which we asked participants to teach a task to a learner robot played by the first author (Section \ref{sec:method}). We conducted a qualitative, bidirectional analysis (Section \ref{sec:task-analysis}) of the data collected to characterize various aspects of actions intended and executed by human teachers when teaching a robot in a situated setting. Our contribution includes a taxonomy that comprises the domain ontology, concept expressions, modalities, and intentions expressed by teachers in this setting.  
Then in Section \ref{sec:discourse-analysis}, we analyze how a computational theory of collaborative interactions - SharedPlans \cite{grosz-kraus, lochbaum-1998-collaborative} can be used to manage human-robot interactions in an ITL setting and support different teaching styles. We conclude in Section \ref{sec:conclusion} with our outlook for future work directed toward developing methods for natural interaction in ITL systems. 
\vspace{-0.2cm}
\section{Related Work}
\vspace{-0.2cm}
Even though robot learning from humans has been studied widely \cite{Thomaz2016ComputationalHRI}, relatively few efforts have looked at how humans would like to teach robots. Some prior work has looked at how people use reinforcement signals such as reward and punishment \cite{knox2012humans, Thomaz2008} to teach robots. Khan et al. \cite{Khan2011} on the other hand, study human teaching strategies, while exploring a single concept, that of ``graspability'' through binary labels assigned to photos of everyday objects.

There is even less work in the domain of teaching complex tasks and procedures. Gil \cite{gil2015} describes the challenges in using human tutorial instruction to teach complex procedures. Gil describes that this can be complex because procedures involve an abundance of interrelated information in terms of situating relevant objects in the environment, steps and substeps of the procedure, as well as conveying control structures. Additionally, human instruction can inherently contain errors and omissions, which one must account for, while designing a natural interface (or in this case, a learning robot). Towards this goal, the contributions in \cite{kaochar2011towards, marge2011towards} examine human tutorial instruction in HRI. Kaochar et al. \cite{kaochar2011towards} conducted a Wizard of Oz experiment to study the various teaching styles exhibited by teachers while teaching a simulated robot a complex task. The teachers could teach via demonstration and examples, provide reinforcement signals and test the agent's skills. They recommend that teaching interfaces must help facilitate teaching using one mode while refining through others for feedback, as well as intermittent testing of the agent's knowledge. Marge et al. \cite{marge2020s} qualitatively analyzed the \textit{Diorama} corpus comprising human-human and human-robot interactions in a collaborative navigation task and created a taxonomy of intentional \textit{dialogue move} types. 
They observed that humans issued direct instructions requesting physical actions to robots; but with other humans, they engaged in plan level conversations. They propose that robots must be capable of translating high-level intentions to realizable physical actions. While both of these works emphasize the need to  understand different aspects of teaching, neither propose concrete recommendations on how a robot can respond appropriately and further the teaching interaction successfully.

Maclellan et al. \cite{maclellan2018framework} propose a Natural Training Interactions (NTI) framework that identifies various teaching patterns in natural human-machine teaching. 
Backed by an established theory SharedPlans, we delve deeper into \emph{direct instruction} and \emph{apprentice learning} teaching patterns described in the NTI framework and propose constructs to study and implement these patterns in ITL robots \cite{breazeal2004teaching, chai2018language, li2020interactive, Mohan2012, mohan2020characterizing}. 


\begin{figure}[!t]
  \centering
   \includegraphics[width=0.8\columnwidth]{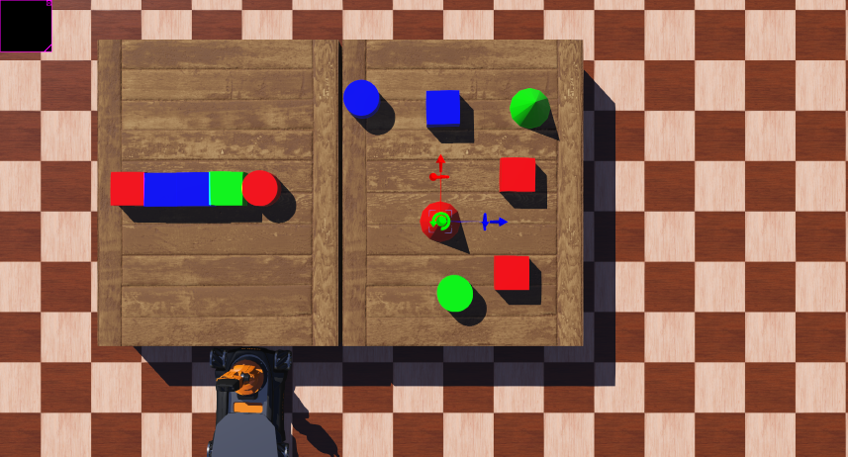}  
  \caption{Example of a wall shown to participants}
  \label{fig:study-env}
  \vspace{-0.55cm}
\end{figure}

\section{Study Design}
\label{sec:method}
\vspace{-0.15cm}
We designed an observational study in which people taught a learning partner through free-form, multi-modal interaction. Due to the ongoing COVID-19 pandemic, we conducted our study remotely using a video-conference application.


\subsection{Study Materials}
\vspace{-0.1cm}
We developed a simulated, robotic, table-top world (shown in Fig. \ref{fig:study-env}) in \href{https://cyberbotics.com/}{Webots}. The world has two tables: the one on the right was designated as the storage area with a set of simple 3-dimensional objects of varying colors and shapes and the one on the left was designated as the main work area where objects could be placed for teaching. We designed our study\footnote{Accessible figure descriptions, teaching transcripts and related study materials can be found at \url{https://git.io/JcWw8}.\label{footnote:1}} as a contextual, semi-structured interview between a researcher, playing the role of the robot (henceforth ResAgent), and the participant playing the role of the teacher. The participants were aware that ResAgent was played by the researcher. The simulated environment was run on the researcher's computer and the participant was given remote access to it. The researcher and the participant observed the same world view presented in Fig. \ref{fig:study-env} and manipulated objects in the world using a mouse pointer. Even though the objects could also be manipulated by a simulated robotic arm, we only used the mouse pointer to allow for more flexibility during requests and action executions.  

We asked the participants to teach ResAgent to build two kinds of walls (a single color wall and a multicolor wall), an example of which is shown in Fig. \ref{fig:study-env}. We chose this task since it could provide participants with a broad range of teaching options that draw on a rich  ontology of domain concepts 
and task knowledge.
We asked participants to use simple sentences and wait for ResAgent's response before moving to the next instruction. We informed  participants that ResAgent did not know anything about the objects in the environment, but that it could learn concepts such as colors, shapes and relations from their instructions. We also informed  participants that ResAgent could point to and move  objects (after it has learned what they are) when asked.

We developed a response protocol for ResAgent to follow while interacting with a participant. This protocol was based on our understanding of learning algorithms implemented in ITL robots as well as some exploratory pilot studies. 
ResAgent simulated learning from a single example or a demonstration. For example, upon being 
told  the color or shape of an object, ResAgent would simulate the learning of the corresponding object label. Similarly, ResAgent would learn to recognize and use relationships such as \emph{next to} 
 from a single example in the domain. Upon successful learning, ResAgent acknowledged its learning by responding (with variations of) \emph{``I have learned the concept.''} 
ResAgent could also explain why a learning failure occurred in specific scenarios. 
For example. if the participant used a concept not taught previously, for example \emph{``right of''} in \emph{``Move the red block to the right of the green block''},  ResAgent would  respond \emph{``I don't know right of''}. If the participant asked ResAgent to perform an action that had not been demonstrated, it would respond with \emph{``I cannot do that''}. Movement actions were assumed to be primitive achievable actions. 
On encountering any unknown action or an instruction that had more than one unknown concept
, ResAgent would respond with \emph{``I don't understand.''}. There were no restrictions on what the teacher could say or what strategy  could be pursued. This was done to provide maximal flexibility to the teacher as well as to elicit different teaching strategies.

\subsection{Conducting the Study}
\vspace{-0.15cm}
The research company's internal review board approved the study materials and protocol as \emph{Exempt}. We recruited participants via internal mailing lists at the research company as well as through personal connections. During recruitment, we were careful to select participants that differed in demographics and backgrounds. We recruited $10$ participants ($5$ female, $4$ male, $1$ non-binary). Their age range distribution was: 18-24: 3, 25-34: 2, 35-44: 3, 45-54: 1, 55-64: 1.

Virtual meetings were scheduled with the participants. We provided participants with a  form that contained the study description and requested consent to record the video, audio and the shared screen manipulation during the study. Each participant's recorded video was later transcribed through an automated transcription service. Due to the remote nature of the study, we encountered some technical issues. Participants sometimes experienced intermittent lag, and were unable to see the environment or move the objects during those times. In those situations, some participants asked the researcher to move objects for them. We also experienced some interruptions due to work-at-home restrictions. However, all participants completed the study task.


\section{Qualitative Bidirectional Analysis}
\label{sec:task-analysis}
\vspace{-0.1cm}
Human teaching is substantially complex but good teaching tends to be structured. We designed our analysis to uncover  structure in human teaching in a situated human robot interaction. To guide our analysis, we introduce a few constructs. Our conjecture is that a human teacher organizes their teaching using a sequence of \textbf{lessons} that comprises setting up a scenario in the environment and providing corresponding information using different modalities. Each lesson is targeted at a \textbf{concept} in the \textbf{domain ontology}. For example, a teacher might introduce a color by pointing to an object and describing it as \emph{red}. During a lesson, the teacher has a variety of \textbf{intentions}, such as defining a concept, introducing an example, or testing the student's knowledge. These can be realized through explicit communicative actions such as commands or questions. Finally, the teacher organizes the sequence of lessons into a \textbf{curriculum} which incrementally introduces various concepts and tasks. In our analysis, we ask the following questions.


\noindent \textbf{1. What are the different types of concepts in the domain ontology and which ones are taught by a human teacher?}
A good understanding  is important to craft appropriate  
algorithms to enable a robot to learn the full space of concepts. \\
\noindent \textbf{2. How do human teachers communicate during a lesson and what modalities do they use to provide information?}
For the robot to learn concepts correctly, it is important that it understands how they are expressed by the teacher (for example, by combining language and gesture in an instruction).   The manner in which the concept is expressed will also influence how the robot adds to or modifies its internal representation of the concept. \\
\textbf{3. What communicative actions are employed by human teachers and what underlying intentions motivate the selection of those actions?}
Answering this question is critical for informing a student robot design  that  is able to: a) recognize the teacher's intentions behind a  communicative action and b) provide a correct response through relevant communicative or physical actions, or by adopting an intention to perform further internal reasoning and learning.\\
\textbf{4. Is there individual variability in the curriculum construction and delivery?} 
People have different sets of assumptions, experience and prior knowledge, which can impact the curriculum that they follow during teaching. We want to understand these variations to build robots that can flexibly adapt to different teachers.

To answer these questions, we conducted a qualitative analysis of the study data. We preprocessed this data to create a dataset that organized the interactions as a sequence of turns. Each turn comprised not only the verbal utterance but also concepts mentioned, gestures used, object configurations, and whether a failure occurred before, during, or after the utterance. This dataset had 1142 total turns comprising 576 participant turns (of which 493 were valid instructions addressing ResAgent and the rest were turns of participants thinking out loud), 458 ResAgent turns and 108 turns that were exchanges between the participant and the researcher discussing technical issues or the study itself.

Our qualitative analysis was bidirectional \textemdash  we conducted both top-down and bottom-up analyses. In our \emph{top-down analysis} \cite{ramaraj2020understanding}, we studied the literature describing ITL robots and general interaction methods in HRI. This included prior work on communication modalities such as language \cite{chai2016collaborative, tellex2014asking, perlmutter2016situated}, gaze \cite{perlmutter2016situated, Thomaz2008}, gestures \cite{perlmutter2016situated, chao2010transparent} and visualization \cite{perlmutter2016situated}. Prior research has also studied concept expressions during teaching such as examples and demonstrations \cite{mohan2020characterizing, alexandrova2014robot} and verbal definitions \cite{kirk2016learning}). We used this analysis to generate {\it a priori} hypotheses about our questions. 

For our \emph{bottom-up analysis}, we conducted an inductive thematic analysis \cite{braun2006using} of our dataset, to both validate our {\it a priori} hypotheses, and extend them. We first manually assigned open codes to each turn in the dataset. For example, P10's instruction of \textit{``Robot, these are green''} was assigned the code \texttt{P10 provides information}. We identified 76 unique participant-specific and 22 unique ResAgent-specific open codes. We then conducted axial coding, a process where different codes across the dataset were clustered together to identify themes. 
Through repeated analysis and identification of different themes, we validated our {\it a priori} hypotheses --- we extracted turns that could be considered as representative of our {\it a priori} hypotheses. We further analyzed the turns that could not be characterized within our {\it a priori} hypotheses and expanded our hypotheses.

\begin{figure}[!t]
  \centering
   \includegraphics[width=0.8\columnwidth]{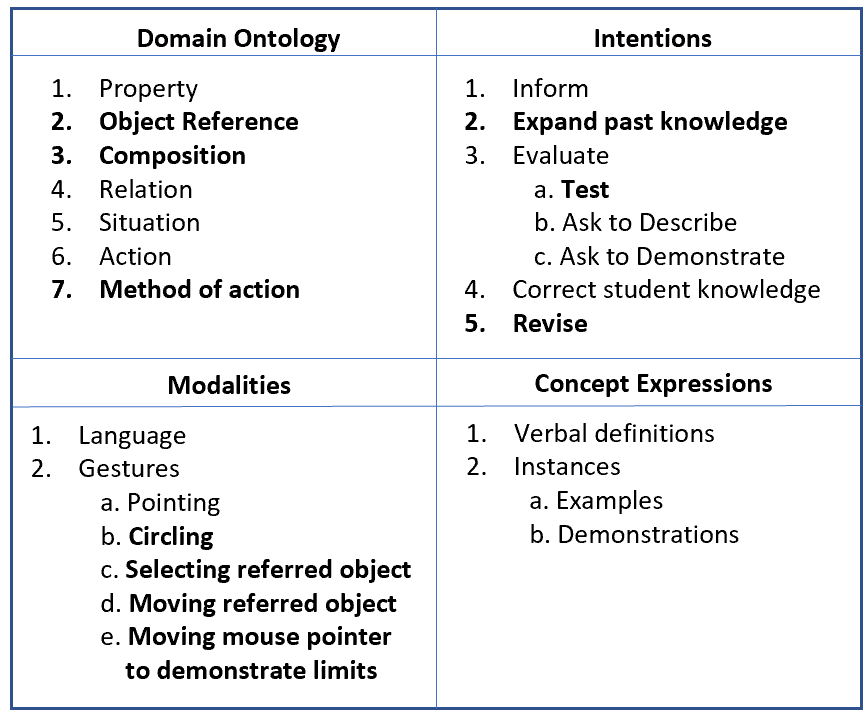}  
  \caption{Taxonomy of different aspects of situated human-robot task teaching generated through our qualitative analysis. The values in \textbf{bold} represent the aspects that emerged through our bottom-up analysis, and the remaining values represent the aspects that were part of our {\it a priori} hypotheses.}
  \label{fig:table-taxonomy}
  \vspace{-0.6cm}
\end{figure}

We generated a comprehensive taxonomy of teaching aspects in a situated human-robot scenario that is presented in Fig. \ref{fig:table-taxonomy}. This taxonomy describes our \textit{a priori} and extended hypotheses, and answers some of the questions that we posed at the beginning of our analysis. It comprises domain ontology (Question 1), types of concept expression, and modalities (Question 2), and intentions (Question 3). We describe these in further detail in the following subsections.
\subsection{Domain Ontology}
\label{sec:domain-ontology}
\vspace{-0.15cm}
In the domain of our analysis, we observed the following types of concepts presented by the teacher:
  \begin{enumerate}
        \item Property: Participants in our study taught the robot how to recognize and label different properties, such as shape and color (e.g., ``These objects are \emph{blue}'').
          \item Relation: 
          Participants either explicitly described a spatial relation between two objects (e.g., ``Robot, the green cone is \emph{left of} the green cube''), or implicitly while demonstrating an action (e.g., ``I am moving the blue cylinder \emph{left of} the red cone''). 
        \item Object Reference: To reference objects, participants either used distinguishing properties (e.g., \emph{red box}) or gestures combined with deictic terms (e.g., ``\emph{This} object is now a part of the wall'').
        \item Composition: Compositions were introduced as complex concepts, composed from basic concepts such as properties, objects, and relations (e.g., ``This is a \emph{wall}'').
        \item Situation: Participants described and referred to 
        the shared environment to provide contextual and spatial knowledge (e.g., ``This point is the \emph{center of the table}''). %
        \item Action: Participant lessons included describing and demonstrating actions, as well as evaluating the robot's ability to perform actions. (e.g., ``\emph{Move} the blue cylinder to the right of the red cone'').
        \item Method of action: When teaching the steps needed to build a wall, it was clear that ``building a wall'' was a high-level action performed ``by'' executing the indicated steps (i.e., the {\it method} for that action). 
    \end{enumerate}

\subsection{Concept Expression and Modalities}
\label{sec:modality}
\vspace{-0.15cm}
Consistent with our top-down analysis, we found that participants used a combination of definitions and instances (examples, demonstrations) to describe and refer to different types of concepts. We present some concrete examples of these in section \ref{sec:taxonomy}. By way of modalities, we observed participants using the mouse to \emph{point} to, \emph{select} and \emph{circle around} the referred objects. Participants also used the mouse to \emph{demonstrate} boundaries of situation, as well as actions. Typically, we observed participants use these gestures along with deictic terms when they referred to \emph{instances} of the domain ontology in the environment. Multiple participants referred to existing objects, or those that were last placed, through language alone. The gestures observed in our study are clearly restricted by the affordances \cite{gibsonaffordance} available in this simulated environment. However, this data supports our hypothesis that teachers naturally use non-verbal modalities along with language for communicative actions. 

\vspace{-0.1cm}
\subsection{Communicative Actions and Intentions}
\label{sec:taxonomy}
\vspace{-0.15cm}
We observed the following intentions and 
 communicative actions  by participants P1--P10.

\textbf{1. Inform}.  The teacher can design a lesson to teach a new concept to a student, by intending to inform in one of three different ways (i.e., methods). The teacher can
\textbf{instantiate} a concept (that is, provide an instance or example of it). 
One of P8's instructions was \emph{``All these objects are cylinders.''} This instruction provides names for concepts present in the shared situation. 
A second method is to \textbf{describe}.  For example, P9 provided a definition of a wall: \emph{``A wall is a line made up of multiple rectangles.''}. Finally, the teacher can inform by using a {\bf hybrid} method. For example, P5 taught ResAgent the property color by saying \emph{``I want the robot to note that this is another color which is blue.''} Through this utterance, P5 provided a conceptual definition that blue is a color, as well as indicated a property about \textit{``this''} on the table.

\textbf{2. Expand past knowledge}.  
A  teacher may continue a lesson about a concept by providing more information about it, thereby expanding the student's past knowledge. 
We saw this in the form of providing \textbf{additional} examples of a property, object or composition. For example, P10 provided ResAgent with multiple wall configurations with the instruction \emph{``This is a wall,''} to help it generalize its understanding of a ``wall.'' Some participants provided more information about a concept in the form of \textbf{distinguishing} information. For example, during P6's first lesson of demonstrating the wall building process, P6 decided to expand ResAgent's knowledge of a block. After using and referring to only "cube-shaped blocks", P6 moved a cylinder to the table and said \emph{``There are some blocks that are not shaped like cubes.''} 
Some participants also provided \textbf{negative} information, presumably so that ResAgent could learn correct concept boundaries using both positive and negative examples. For example, in P4's lesson about a ``blue wall'', P4 first built a wall and said to ResAgent,  \emph{``This is a blue wall.''} Once ResAgent confirmed that it had learned successfully, P4 scattered the objects across the table and said \emph{``This is not a blue wall.''} 

\textbf{3.  Evaluate}.  We often found that the participant tried (that is, intended  \cite{ortiz-commonsense-cause}) to {\it evaluate} ResAgent's progress as a part of a lesson 
by eliciting information from ResAgent to assess its state of knowledge. We found that they accomplished this through the following methods. In a \textbf{test} method, the teacher can evaluate a robot's knowledge by asking yes/no questions to verify its knowledge of concepts and actions. 
For example, 
P4 during their lesson about a ``wall'', built a wall and asked ResAgent \emph{``Is this a wall?''} to verify that it had learned this composition correctly. In a \textbf{describe} method, the teacher can evaluate whether the robot can describe its conceptual knowledge and shared environment, or retrospect on its experience and use its conceptual knowledge to summarize it. 
An example is seen when P10 asked ResAgent to describe the result of its actions: \emph{``Can you tell me what you've built?''}. When the robot either stops arbitrarily or indicates that it is unable to progress any further, the teacher can ask the robot to \textbf{explain}.
During P9's lesson about a wall, after providing examples, P9 asked ResAgent to build a wall. However, ResAgent did not have enough knowledge to execute the task and responded \emph{``I cannot do that.''} P9 then asked \emph{ ``Why can't you do that?''} in response. ResAgent was unable to answer this question since it was not part of the response protocol and just repeated \emph{``I cannot do that.''} 
It would be useful for a robot to provide a response that helps the teacher identify the missing concept, and design an explicit lesson to teach it, so that the robot can recover from this lack of knowledge.
Lastly, a teacher can ask a robot to \textbf{demonstrate} an action to confirm whether the robot has learned the correct procedural {\it how-to} knowledge. Multiple participants instructed ResAgent to ``build'' or ``make'' a wall at the end of their lessons to evaluate its learning. \\
\indent \textbf{4.  Correct student knowledge}. Learning failures can occur when the robot does not have the requisite or correct concept definitions or when its concept definitions are either over-general or over-specific; this can lead to incorrect scene understanding or task execution. During a lesson, when a participant used a concept that they had not taught yet or discovered during an \textit{Evaluate} action that ResAgent did not know a concept, 
they either provided definitions or demonstrated examples of the concept 
to update ResAgent's knowledge. For example, when P10 asked ResAgent to perform an action by saying  \emph{``Robot, can you move the green cylinder immediately to the left of the green cube?,''} ResAgent responded \emph{``I don't know left of.''} P10 then 
\emph{corrected} by saying \emph{``The green cone is left of the green cube,''} and pointing to the objects while referring to them. \\
\indent \textbf{5.  Revise instruction}.   In line with the functional reduction strategies described in \cite{gieselmann2006comparing}, we observed participants revising their instructions by either self-correcting when they made mistakes during an instruction, or rephrasing a recent instruction to make it clearer when ResAgent did not understand. During P6's lesson about compositions, P6 said \emph{``A vertical line is when two or more blocks are above and below one another and like a Northwest orientation.''} When ResAgent said it did not understand, P6 \emph{revised} their instruction and instead demonstrated the concept by saying \emph{``So this line is horizontal and this line is vertical.''} P1, during their lesson introducing domain ontology, asked ResAgent to perform an action, but immediately followed it up with \emph{``Cancel that.''} This case is challenging for the robot as it has to determine which information, if any, from its previous interaction needs to be revised or eliminated.

\begin{figure}[!t]
  \centering
  \includegraphics[width=\columnwidth]{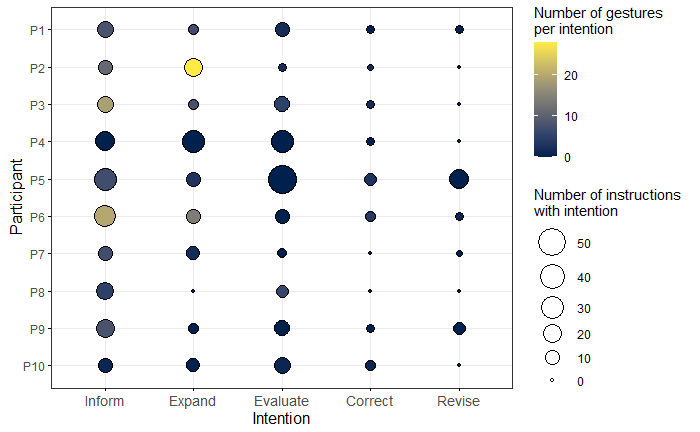}  
  \caption{This color-intensity balloon plot depicts for each participant, the total number of instructions with each intention, and the total number of gestures used during instructions with a specific intention. Bigger circles correspond to more instructions, whereas lighter color intensity corresponds to higher number of gestures.}
  \label{fig:balloon-plot}
  \vspace{-0.5cm}
\end{figure}


\subsection{Variability in Curriculum Construction}
\label{sec:curriculum-analysis}
\vspace{-0.15cm}
Finally, as an answer to Question 4, Fig. \ref{fig:balloon-plot} illustrates the the differences in teaching styles of the study participants. Despite teaching the same task, the lengths of teacher-student interaction varied across participants. P7 taught the task in 46 turns while P5 took longer and taught the task in 223 turns. We also observed people express different teaching intentions to varying degrees. We see that some participants (P1, P3, P8) structured their teaching primarily using \emph{inform} and \emph{evaluate} intentions. Others (P2, P4, P5) \emph{expanded} previously introduced concepts. We found that relatively few participants used \emph{correct} and \emph{revise} intentions. However, P9 and especially, P5 frequently \emph{revised} their instructions when ResAgent did not understand. Finally, we found that all participants except P4 used gestures to teach but not to the same degree. Some participants (P2, P3, P6) used gestures extensively while others were conservative. The concentration of gestures in \emph{inform} and \emph{expand} intentions suggests that gestures are useful in providing extra-linguistic information to learn from.

\begin{figure}[t!]
\centering
\subfloat[Curriculum employed by P6\label{fig:sub-flow-P6}]{\fbox{\includegraphics[width=0.96\columnwidth]{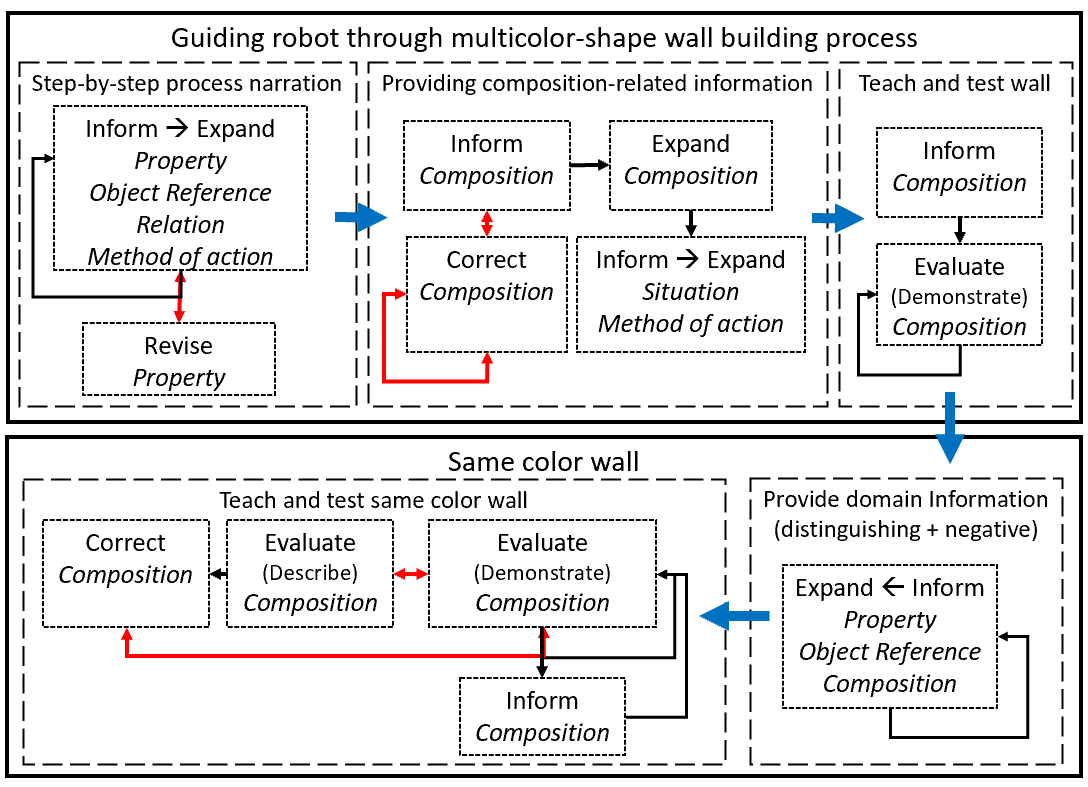}}} \qquad
\subfloat[Curriculum employed by P10\label{fig:sub-flow-P10}]{\fbox{\includegraphics[width=0.96\columnwidth]{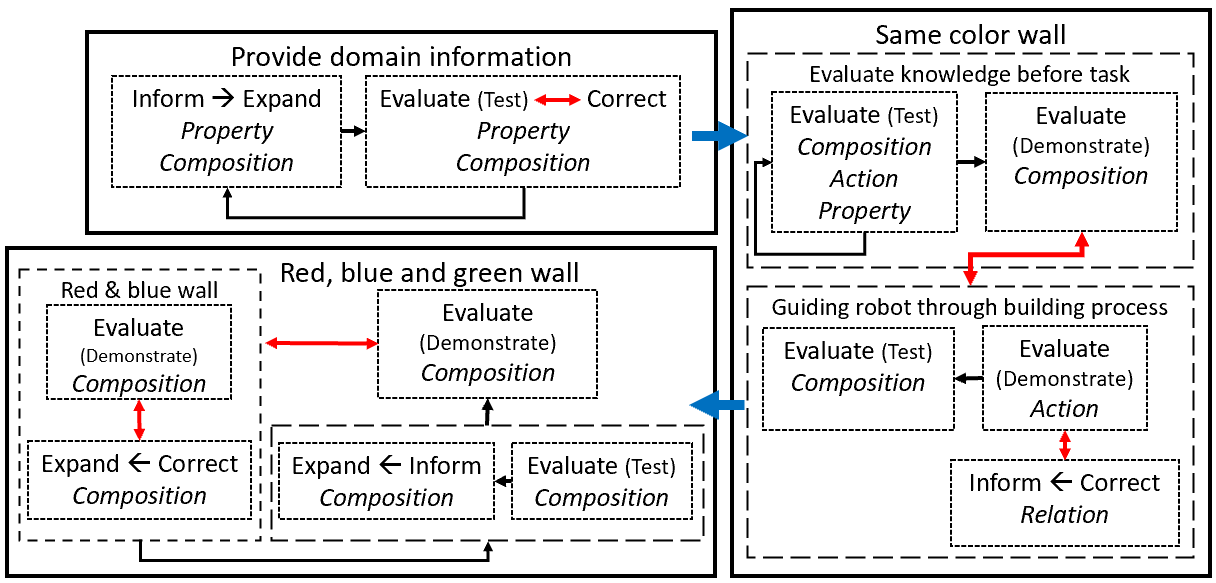}}} 
\caption{Flowcharts of the P6 and P10 curricula. Innermost boxes with dotted borders contain the intended communicative actions that P6 and P10 executed, along with the referred domain ontology. Bounding boxes with dashed borders represent the lessons formed by grouping of actions for a given purpose (present at the top of the box). Black arrows depict the flow of communicative actions, and the blue arrows depict the flow of lessons. The red double-headed arrows refer to the sub-lesson that the teacher provided, when an instruction resulted in ResAgent's failure, before proceeding ahead.}
 \label{fig:instruction-flow}
 \vspace{-0.6cm}
\end{figure}

Fig. \ref{fig:instruction-flow} examines in  more detail the differences in the curricula employed by two participants P6 and P10. Fig. \ref{fig:sub-flow-P6} and \ref{fig:sub-flow-P10} illustrate how P6 and P10 sequenced their communicative actions to form lessons that constituted the curricula\footnote{P6 described their background as: \emph{"I have a degree in Computer Engineering, emphasis in Robotics. Pursuing a PhD in Human Computer Interaction, have some knowledge of Human Robot Interaction \& research."} P10 described their background as: \emph{``I am a PhD student in English Language and Literature, and I have no background in CS, computer programming, or robotics.''} \label{footnote:2}}.  
P10 pursued a bottom-up interactive teaching strategy. P10 began by presenting examples from the domain to teach concepts. This can be observed in P10's first lesson, where P10 used \emph{inform} and \emph{expand} actions to provide examples of properties (colors, shapes) and compositions (wall) to ResAgent. P10 then \emph{tested} ResAgent's ability to identify these concept instances in the domain and \emph{corrected its knowledge} through examples when it failed. P10's next lesson was to teach ResAgent the process of building the same color wall, and introduced relations through \emph{inform} and \emph{correct} actions. After ResAgent successfully \emph{demonstrated} building a wall, P10 proceeded to the lesson about multicolor walls. P10 provided examples of walls through \emph{inform} and \emph{expand} actions, and \emph{corrected} ResAgent's knowledge when it could not identify them. P10's instruction ended when ResAgent successfully \emph{demonstrated} building the \emph{red, blue and green wall}.

P6 on the other hand, pursued a top-down narration-type strategy. P6's first set of lessons involved describing and demonstrating the process of building a wall, that we explore further in section \ref{sec:discourse-analysis} while discussing Fig. \ref{Figure-scenario}. P6's second lesson began with using \emph{inform} and \emph{expand} actions to provide domain information, in preparation to teach same color walls. P6 decided to first \emph{test} ResAgent's ability to build a ``red'' wall. When it could not do it, P6 used the \emph{inform} action to provide it an example. P6 repeated this process with other color walls, and ended their second lesson when ResAgent successfully \emph{demonstrated} building a same color wall. 
\section{SharedPlans in ITL Robots}
\label{sec:discourse-analysis}
\vspace{-0.15cm}
Our analyses in the previous section demonstrate that human teaching is intentional, dynamic and has a high degree of individual variability. Teachers form intentions to assess and improve the student's domain knowledge. The intentions are expressed through a repertoire of communicative \emph{teaching}  actions. To exploit the full range of information available in human teaching, an ITL robot must be able to sustain a flexible interaction with the teacher. While an implemented computational model is not within the scope of this paper, below we discuss how a particularly well-developed theory of collaboration called SharedPlans (SP) \cite{grosz-kraus, lochbaum-1998-collaborative, ortiz-dial}  can be used to guide the design of interaction models in ITL robots. 

SP makes use of a number of formal constructs that enable one to model a collaboration. 
The first important element is an \textbf{intention}.  There are two types of intentions:  an {\it intention-that} ranges over participants and  states of the world, including the state of teaching.  They capture a participant's commitment to a  world state in which some fact holds. For example, the teacher and the student might have an \emph{intention-that} the student understand what a wall is. The second type is called an {\it intention-to} that ranges over participants and actions (communicative, physical, and internal), and represents the commitment that a participant has to performing some action in the future. For example, a student may \emph{intend-to} convey its success in understanding the concept of a   wall  to the teacher. An ITL robot must be encoded with methods that allow it to infer (that is, recognize) the overarching intentions of the teacher so that it can act in a way that advances rather than inhibits the achievement of a recognized intention. 

\begin{figure}[t!]
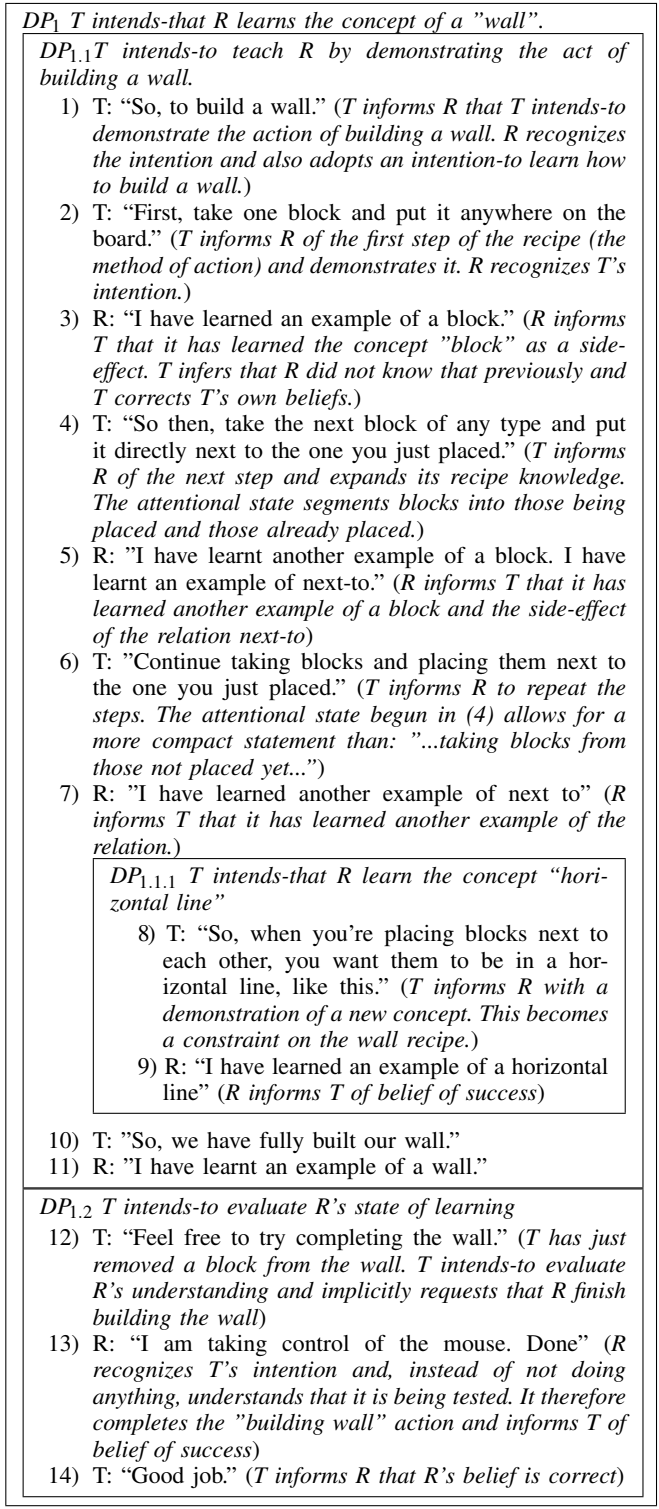

\vspace{.1in}
\small 	
  \fbox{
 	\begin{minipage}{0.95\columnwidth}
\raggedright $DP_{1}$ {\it  T intends-that R learns the concept of a "wall".} 

	\fbox{
		\begin{minipage}{0.95\columnwidth}
     $DP_{1.1}${\it T intends-to teach R by demonstrating the act of building a wall.}\\
     \begin{enumerate}
\vspace{-0.15in}
	\item T: ``So, to build a wall.'' ({\it T informs R that T intends-to demonstrate the action of building a wall. R recognizes the intention and also adopts an intention-to learn how to build a wall.})
	\item T: ``First, take one block and put it anywhere on the board."  ({\it T informs R of the first step of the recipe (the method of action) and demonstrates it.   R recognizes T's intention.})
	\item R: ``I have learned an example of a block."  ({\it R informs T that it has learned the concept "block" as a side-effect.   T infers that R did not know that previously and T corrects T's own beliefs.})
	\item T: ``So then, take the next block of any type and put it directly next to the one you just placed.''  ({\it T informs R of the next step and expands its recipe knowledge.   The attentional state segments blocks into those being placed and those already placed.}) 
	\item R: "I have learnt another example of a block.  I have learnt an example of next-to." ({\it R informs T that it has learned another example of a block and the side-effect of the relation next-to})
	\item T: "Continue taking blocks and placing them next to the one you just placed." ({\it T informs R to repeat the steps.  The attentional state begun in (4) allows for a more compact statement than: "...taking blocks from those not placed yet..."})
	\item R: "I have learned another example of next to" ({\it R informs T that it has learned another example of the relation.})

\fbox{
	\begin{minipage}{0.85\columnwidth}
$DP_{1.1.1}$ {\it T intends-that R learn the concept ``horizontal line''}
      \begin{enumerate}[label=\arabic*]

   \setcounter{enumii}{7}
  
        \item  
        \hspace{-.08in}) T: ``So, when you're placing blocks next to each other, you want them to be in a horizontal line, like this." ({\it T informs  R with a demonstration of a new concept.   This becomes a constraint on the wall recipe.})
     	\item 
      	\hspace{-.07in}) R: ``I have learned an example of a horizontal line'' ({\it R informs T of belief of success}) 
      	\end{enumerate}
	\end{minipage}
	}
	\vspace{.04in}
	\setcounter{enumi}{9}
	\item T: "So, we have fully built our wall." 
	\item R: "I have learnt an example of a wall."
\end{enumerate}
      		\end{minipage}
		}

\fbox{
	\begin{minipage}{0.95\columnwidth}
	$DP_{1.2}$ {\it T intends-to evaluate R's state of learning}
	\begin{enumerate}
	\setcounter{enumi}{11}
	\item T: ``Feel free to try completing the wall.'' ({\it T has just removed a block from the wall.  T intends-to evaluate R's understanding and implicitly requests that R finish building the wall}) 
	\item R: ``I am taking control of the mouse. Done'' ({\it R recognizes T's intention and, instead of not doing anything, understands that it is being tested.  It therefore completes the "building wall" action and informs T of belief of success})
	\item T: ``Good job.'' ({\it T informs R that R's belief is correct})
	\end{enumerate}
	\end{minipage}
	}
\end{minipage}
}

\caption{Actual interaction between the participant teacher P6 (T), and the robot student ResAgent (R) that illustrates  SP constructs in the context of ITL. T uses demonstration and narration to teach the concept of a wall. The anonymized video of this interaction can be found at \url{https://youtu.be/bspbZhsCOpw}.}
\label{Figure-scenario}
\normalsize    
\vspace{-0.6cm}
\end{figure}

The next element is that of a {\bf belief}. Beliefs 
capture a participant's view of the world, in some symbolic representation.
Participants' beliefs can differ, as they do at certain points in   ITL settings: the recognition of such differences often triggers a teaching action to correct a student's beliefs.  Sometimes, the teacher must introduce a new concept before the student can form an appropriate belief that can be correctly tied to its sensory perceptions: for example, knowledge of which structures  constitute a  wall.
Successful instruction also requires that participants acquire \emph{mutual belief} of the current state of the world and interaction; this is facilitated through a perceptually obtained common ground.  

Since, a major focus of SP is on tasks and actions, the theory introduces {\bf recipes} that reflect action decomposition.  For example, the task of building  a wall can be hierarchically decomposed into a recipe or set of steps, such as picking up and placing the first block, picking up the next block and placing it next to the first one, and so on. Each sub-action in a recipe can be viewed as part of the {\it method} for the higher level action. Similar to beliefs, a robot's recipes may be incomplete or incorrect. The teacher helps the robot learn the right recipes to successfully perform actions in the domain. 
Since teacher communications are also actions, they too can be decomposed into more basic actions through recipes. A teacher will often sequence  teaching actions in a way that reflects a personal teaching strategy. 
During an interaction, the robot has to infer the most likely  intention that the  teacher has adopted given its library of recipes.   For example, if the teacher corrects a robot misconception, that inform action is grounded in the teacher intention-that the robot modify its beliefs accordingly. The robot must recognize that implicit intention, and ask for clarification if it cannot.

Fig. \ref{Figure-scenario} illustrates  SP constructs, as applied to task-based dialogue processing, in an annotation of   a portion of P6's teacher-robot interaction. Associated with groups of utterances (shown inside a box) is a {\it discourse purpose} (DP). The sub-boxes in the figure indicate that the enclosed discourse segment has its own purpose and is subsidiary (that is, contributes in some way, to the higher level purpose). The grouping of {\it utterances} into segments, each of which serves some particular role in the discourse, is referred to as the {\bf linguistic structure}. 
A second, deeper structure is referred to as the {\bf intentional structure}. This consists of the DP of each segment, expressed in terms of the language of beliefs and intentions discussed earlier, together with relations between another DP.   
For example, in Fig. \ref{Figure-scenario},  utterances 1-11 describe {\it how} to build a wall; i.e., the recipe for building a wall. The sub-segment consisting of utterances 8 and 9 is a sub-dialogue describing a  horizontal line constraint on the positioning of the block.  In $DP_{1.1.1}$,  T intends-that R learn the concept ``horizontal line'' which contributes to the success of the higher level $DP_{1.1}$ and induces a structure between those two segments.   The DPs can be thought of as related in the same way that a lower level action is related to a higher level action in a recipe.   
The final important element of dialog is referred to as the {\bf attentional state}: at any stage of an interaction, the participants access only a subset of salient entities under discussion. One example of the evolution of the attentional state is illustrated in utterances (4) and (6). Since teacher-student interactions are situated, entities are often brought into salience through pointing and deictics.

The SP framework can address many of the requirements that we have uncovered while attempting to answer the questions that we posed in Section \ref{sec:task-analysis}: 

{\bf 1.}  SP provides a rich representation of actions and their decomposition together with a non-propositional logical language that supports representation of properties and relations.

{\bf 2.} Dialogue structure is not stipulated ahead of time but can follow the natural improvisational flow of normal conversations.  Helpful actions are automatically triggered by intentions-that so that  
perceived teaching/learning obstacles can be overcome.   Alternative modalities can simply be represented as special actions with their own effects.  

{\bf 3.} SP supports multiple intentions and since teaching intentions are grounded in action they can be decomposed like any other action.   Since intentions are closely tied to action, explanations for why a student did something can be easily generated.  Partiality of intentions and beliefs naturally support {\it expand} and {\it revise} \cite{Ortiz-DIS}.

{\bf 4.} Individual variability of instruction and curriculum is directly supported in SP as shown above: DPs can follow whatever order a teacher finds suitable.

\section{Discussion and Conclusion}
\label{sec:conclusion}
\vspace{-0.15cm}
In this paper, we study how humans teach to design ITL robots that can participate in a natural teaching interaction. Towards this end, we designed and conducted a study in which participants taught a learning agent in a simulated, robotic domain. Our analysis shows that people naturally decompose a complex task into a variety of concepts (properties, relations, compositions, etc.) and incrementally introduce these as parts of lessons in a curriculum. In a lesson,  conceptual knowledge is presented through definitions, demonstrations and examples, and use a combination of language and non-verbal modalities to convey information. Teachers adopt  a variety of intentions in a lesson to introduce concepts as well as to understand and evaluate a student's competency. We propose a taxonomy that organizes these aspects of human-robot teaching to analyze the teaching process of two participants, thereby demonstrating the utility of the taxonomy in modeling human teaching. 

Our analysis also reveals that there is significant individual variability in how people design and deliver a curriculum. To design a robot that can learn from the full range of information available in human teaching, it is necessary to develop a computational model that can handle this inherent variability. The paper explores the applicability of a computational theory of collaboration, SharedPlans, in managing human-robot teaching interactions. One of the advantages of SharedPlans is its flexibility in adapting its theoretical constructs to many alternative teaching styles by modeling the teaching interaction as a collaboration between the teacher and the student towards a mutual goal. It represents a promising approach to designing a robot that can accommodate different teachers and their teaching strategies. 

While the analyses in this paper are promising, they represent  only an initial exploration directed towards understanding human teaching in  ITL settings. Our study had 10 participants who taught a simple task with a limited set of objects. We expect that other aspects of teaching will emerge with more complex tasks as well as with participants from diverse backgrounds, cultures, and experiences. Another limitation is the remote nature of our study. While we expect the taxonomy of intentions to be applicable in real human-robot interaction scenarios, the experimental setup that we adopted limits the gestures that people can use. Despite these limitations, our work lays the foundation for a structured understanding of ITL teaching  and provides a framework that can be used for building and evaluating future ITL robots.

Often, HRI and, specifically, ITL have approached interactions from a robot-learner perspective by building simple and scripted interaction models. In the future, we will apply the SharedPlans theory to implement a teacher-centered interaction model in an ITL robot that will enable it to adopt correct intentions in response to a teacher's communicative acts, and respond correctly to further the interaction.
\vspace{-0.1cm}
\section*{Acknowledgments}
\noindent We would like to thank Kent Evans for help setting up the study environment and Matthew Klenk for his valuable inputs. We also thank the reviewers for their feedback; and finally, our study participants for their help. This work was begun while the first author was an intern at PARC. 
\vspace{-0.2cm}
\bibliographystyle{IEEEtran}
\bibliography{references}

\begin{thebibliography}{10}
\providecommand{\url}[1]{#1}
\csname url@rmstyle\endcsname
\providecommand{\newblock}{\relax}
\providecommand{\bibinfo}[2]{#2}
\providecommand\BIBentrySTDinterwordspacing{\spaceskip=0pt\relax}
\providecommand\BIBentryALTinterwordstretchfactor{4}
\providecommand\BIBentryALTinterwordspacing{\spaceskip=\fontdimen2\font plus
\BIBentryALTinterwordstretchfactor\fontdimen3\font minus
  \fontdimen4\font\relax}
\providecommand\BIBforeignlanguage[2]{{%
\expandafter\ifx\csname l@#1\endcsname\relax
\typeout{** WARNING: IEEEtran.bst: No hyphenation pattern has been}%
\typeout{** loaded for the language `#1'. Using the pattern for}%
\typeout{** the default language instead.}%
\else
\language=\csname l@#1\endcsname
\fi
#2}}

\bibitem{laird2017interactive}
J.~Laird, K.~Gluck, J.~Anderson, K.~Forbus, O.~C. Jenkins, C.~Lebiere,
  D.~Salvucci, M.~Scheutz, A.~Thomaz, G.~Trafton, R.~Wray, S.~Mohan, and
  J.~Kirk, ``Interactive task learning,'' \emph{IEEE Intelligent Systems},
  vol.~32, no.~4, pp. 6--21, 2017.

\bibitem{norman1983some}
D.~A. Norman, ``Some observations on mental models,'' \emph{Mental models},
  vol.~7, no. 112, pp. 7--14, 1983.

\bibitem{grosz-kraus}
B.~J. Grosz and S.~Kraus, ``Collaborative plans for complex group action,''
  \emph{Artificial Intelligence}, vol.~86, no.~2, pp. 269--357, 1996.

\bibitem{ortiz-dial}
C.~Ortiz and B.~Grosz, ``Interpreting information requests in context a
  collaborative web interface for distance learning,'' \emph{Autonomous Agents
  and Multi-Agent Systems}, vol.~5, pp. 429--465, 2002.

\bibitem{breazeal2004teaching}
C.~Breazeal, G.~Hoffman, and A.~Thomaz, ``Teaching and working with robots as a
  collaboration,'' in \emph{AAMAS}, 2004, pp. 1030--1037.

\bibitem{Mohan2012}
S.~Mohan, A.~H. Mininger, J.~R. Kirk, and J.~E. Laird, ``Acquiring grounded
  representations of words with situated interactive instruction,'' in
  \emph{Advances in Cognitive Systems}, vol.~2, 2012, pp. 113--130.

\bibitem{mohseni2015}
A.~Mohseni-Kabir, C.~Rich, S.~Chernova, C.~L. Sidner, and D.~Miller,
  ``Interactive hierarchical task learning from a single demonstration,'' in
  \emph{HRI}, 2015, p. 205–212.

\bibitem{lochbaum-1998-collaborative}
K.~E. Lochbaum, ``A collaborative planning model of intentional structure,''
  \emph{Computational Linguistics}, vol.~24, pp. 525--572, 1998.

\bibitem{Thomaz2016ComputationalHRI}
A.~Thomaz, G.~Hoffman, and M.~Cakmak, ``Computational human-robot
  interaction,'' \emph{Found. Trends Robotics}, vol.~4, pp. 105--223, 2016.

\bibitem{knox2012humans}
W.~B. Knox, B.~D. Glass, B.~C. Love, W.~T. Maddox, and P.~Stone, ``How humans
  teach agents,'' \emph{IJSR}, vol.~4, no.~4, pp. 409--421, 2012.

\bibitem{Thomaz2008}
A.~L. Thomaz and C.~Breazeal, ``Teachable robots: Understanding human teaching
  behavior to build more effective robot learners,'' \emph{Artificial
  Intelligence}, vol. 172, no. 6-7, pp. 716--737, 2008.

\bibitem{Khan2011}
F.~Khan, X.~Zhu, and B.~Mutlu, ``{How do humans teach: On curriculum learning
  and teaching dimension},'' in \emph{NeurIPS}, 2011.

\bibitem{gil2015}
Y.~Gil, ``Human tutorial instruction in the raw,'' \emph{ACM Transactions on
  Interactive Intelligent Systems (TiiS)}, vol.~5, no.~1, pp. 1--29, 2015.

\bibitem{kaochar2011towards}
T.~Kaochar, R.~T. Peralta, C.~T. Morrison, I.~R. Fasel, T.~J. Walsh, and P.~R.
  Cohen, ``Towards understanding how humans teach robots,'' in
  \emph{International Conference on User Modeling, Adaptation, and
  Personalization}.\hskip 1em plus 0.5em minus 0.4em\relax Springer, 2011, pp.
  347--352.

\bibitem{marge2011towards}
M.~Marge and A.~I. Rudnicky, ``Towards overcoming miscommunication in situated
  dialogue by asking questions.'' in \emph{AAAI Fall Symposium: Building
  Representations of Common Ground with Intelligent Agents}, 2011.

\bibitem{marge2020s}
M.~Marge, F.~Gervits, G.~Briggs, M.~Scheutz, and A.~Roque, ``Let's do that
  first! a comparative analysis of instruction-giving in human-human and
  human-robot situated dialogue,'' in \emph{Proc. of SemDial}, 2020.

\bibitem{maclellan2018framework}
C.~J. MacLellan, E.~Harpstead, R.~P. Marinier~III, and K.~R. Koedinger, ``A
  framework for natural cognitive system training interactions,''
  \emph{Advances in Cognitive Systems}, vol.~6, pp. 1--16, 2018.

\bibitem{chai2018language}
J.~Y. Chai, Q.~Gao, L.~She, S.~Yang, S.~Saba-Sadiya, and G.~Xu, ``Language to
  action: Towards interactive task learning with physical agents.'' in
  \emph{IJCAI}, 2018, pp. 2--9.

\bibitem{li2020interactive}
T.~J.-J. Li, T.~Mitchell, and B.~Myers, ``Interactive task learning from
  gui-grounded natural language instructions and demonstrations,'' in
  \emph{ACL}, 2020, pp. 215--223.

\bibitem{mohan2020characterizing}
S.~Mohan, M.~Klenk, M.~Shreve, K.~Evans, A.~Ang, and J.~Maxwell,
  ``Characterizing an analogical concept memory for newellian cognitive
  architectures implementing the common model of cognition,'' in \emph{8th
  Annual Conference on Advances in Cognitive Systems}, 2020.

\bibitem{ramaraj2020understanding}
P.~Ramaraj, M.~Klenk, and S.~Mohan, ``Understanding intentions in human
  teaching to design interactive task learning robots,'' in \emph{RSS 2020
  Workshop: AI \& Its Alternatives in Assistive \& Collaborative Robotics:
  Decoding Intent}, 2020.

\bibitem{chai2016collaborative}
J.~Y. Chai, R.~Fang, C.~Liu, and L.~She, ``Collaborative language grounding
  toward situated human-robot dialogue.'' \emph{AI Magazine}, vol.~37, no.~4,
  2016.

\bibitem{tellex2014asking}
S.~Tellex, R.~A. Knepper, A.~Li, D.~Rus, and N.~Roy, ``Asking for help using
  inverse semantics,'' in \emph{Robotics: Science and Systems}, 2014.

\bibitem{perlmutter2016situated}
L.~Perlmutter, E.~Kernfeld, and M.~Cakmak, ``Situated language understanding
  with human-like and visualization-based transparency.'' in \emph{Robotics:
  Science and Systems}, 2016.

\bibitem{chao2010transparent}
C.~Chao, M.~Cakmak, and A.~L. Thomaz, ``Transparent active learning for
  robots,'' in \emph{HRI}, 2010, pp. 317--324.

\bibitem{alexandrova2014robot}
S.~Alexandrova, M.~Cakmak, K.~Hsiao, and L.~Takayama, ``Robot programming by
  demonstration with interactive action visualizations,'' in \emph{Robotics:
  science and systems}, 2014.

\bibitem{kirk2016learning}
J.~R. Kirk and J.~E. Laird, ``Learning general and efficient representations of
  novel games through interactive instruction,'' \emph{Advances in Cognitive
  Systems}, vol.~4, 2016.

\bibitem{braun2006using}
V.~Braun and V.~Clarke, ``Using thematic analysis in psychology,''
  \emph{Qualitative research in psychology}, vol.~3, no.~2, pp. 77--101, 2006.

\bibitem{gibsonaffordance}
J.~J. Gibson, ``The theory of affordances,'' in \emph{The Ecological Approach
  to Visual Perception}.\hskip 1em plus 0.5em minus 0.4em\relax Psychology
  Press, 2014, ch.~8, pp. 119--136.

\bibitem{ortiz-commonsense-cause}
C.~L. Ortiz, ``A commonsense language for reasoning about causation and
  rational action,'' \emph{Artificial Intelligence}, vol. 111, pp. 73--130,
  1999.

\bibitem{gieselmann2006comparing}
P.~Gieselmann, ``Comparing error-handling strategies in human-human and
  human-robot dialogues,'' in \emph{Proc. KONVENS}, 2006, pp. 24--31.

\bibitem{Ortiz-DIS}
C.~L. Ortiz~Jr and L.~Hunsberger, ``On the revision of dynamic intention
  structures,'' in \emph{Commonsense Reasoning Symposium}, 2013.

\end{thebibliography}

\end{document}